# Relation Extraction Using Large Language Models: A Case Study on Acupuncture Point Locations

Yiming Li, MS[1], Xueqing Peng, PhD[2], Jianfu Li, PhD[3], Xu Zuo, PhD[1], Suyuan Peng, PhD[4], Donghong Pei, PhD[5], Cui Tao, PhD[3], Hua Xu, PhD[2], Na Hong, PhD[2]

[1]McWilliams School of Biomedical Informatics, The University of Texas Health Science Center at Houston, Houston, TX 77030, USA; [2]Section of Biomedical Informatics and Data Science, School of Medicine, Yale University, New Haven, CT 06510, USA; [3]Department of Artificial Intelligence and Informatics, Mayo Clinic, Jacksonville, FL 32224, USA; [4]Institute of Information on Traditional Chinese Medicine, China Academy of Chinese Medical Sciences, Beijing, 100010, China; [5]The University of Texas MD Anderson Cancer Center, Houston, TX 77030, USA

Corresponding Author: Na Hong (na.hong@yale.edu)

## ABSTRACT

### Objective

In acupuncture therapy, the accurate location of acupoints is essential for its effectiveness. The advanced language understanding capabilities of large language models (LLMs) like Generative Pre-trained Transformers (GPT) present a significant opportunity for extracting relations related to acupoint locations from textual knowledge sources. This study aims to compare the performance of GPT with traditional deep learning models (Long Short-Term Memory (LSTM) and Bidirectional Encoder Representations from Transformers for Biomedical Text Mining (BioBERT)) in extracting acupoint-related location relations and assess the impact of pretraining and fine-tuning on GPT's performance.

**Materials and Methods**

We utilized the World Health Organization Standard Acupuncture Point Locations in the Western Pacific Region (WHO Standard) as our corpus, which consists of descriptions of 361 acupoints. Five types of relations ('direction_of,' 'distance_of,' 'part_of,' 'near_acupoint,' and 'located_near') (n= 3,174) between acupoints were annotated. Five models were compared: BioBERT, LSTM, pre-trained GPT-3.5, fine-tuned GPT-3.5, as well as pre-trained GPT-4. Performance metrics included micro-average exact match precision, recall, and F1 scores.

**Results**

Our results demonstrate that fine-tuned GPT-3.5 consistently outperformed other models in F1 scores across all relation types. Overall, it achieved the highest micro-average F1 score of 0.92.

**Conclusion**

This study underscores the effectiveness of LLMs like GPT in extracting relations related to acupoint locations, with implications for accurately modeling acupuncture knowledge and promoting standard implementation in acupuncture training and practice. The findings also contribute to advancing informatics applications in traditional and complementary medicine, showcasing the potential of LLMs in natural language processing.

# INTRODUCTION

Acupuncture, originating from ancient Chinese medicine, has a history dating back thousands of years [1,2]. It was introduced to the United States in the 1970s, gaining gradual acceptance within the medical community for its therapeutic effects [3,4]. According to the World Health Organization 2019 report, acupuncture is the most widely used traditional and complementary medicine, practiced in 113 out of 120 countries [5].

Acupuncture involves the insertion of thin needles into specific points on the body known as acupoints, which are believed to correspond to channels that conduct Qi, or vital energy [6,7]. Clinical studies have shown acupuncture to be effective in treating a variety of conditions, including chronic pain, migraines, and osteoarthritis [8]. It is also used to alleviate symptoms associated with chemotherapy, such as nausea and vomiting [9]. The World Health Organization recognizes acupuncture as a viable treatment option for a range of disorders, including respiratory, digestive, and neurological conditions [10–12].

The rationale behind acupuncture lies in its emphasis on the precise location of acupoints [13,14]. These points are located along meridians, or energy pathways, where Qi flows [15]. In practice, the accurate location of acupoints is essential to ensure the proper flow of Qi, and acupuncturists need to understand the anatomical structures of the body to accurately locate acupuncture points and ensure the safety and effectiveness of acupuncture procedures [16]. Moreover, acupuncture localization also adheres to the concept of "body cun"(同身寸), where acupuncture points are determined based on certain lengths on the patient's body surface. Acupuncture therapy does not rely on fixed physical dimensions but adjusts according to the patient's body characteristics. While some acupoints are easily identifiable, others may be more challenging to locate, especially for beginners [17]. Acupuncturists undergo extensive training to master the palpation techniques and anatomical knowledge necessary for accurate point location. Several studies suggested that the accuracy and precision vary when locating acupuncture points using different methods [18]. Given the variance in treatment outcomes among these point location techniques, it is essential to leverage informatics technologies to structuralize and computerize the most critical point location knowledge for assisting acupuncture training and practice.

While there are numerous studies on relation extraction (RE) from biomedical literature, none have yet delved into relation extraction from textual acupuncture knowledge. He et al. used the CHEMPROT dataset from BioCreative VI and the DDI dataset to develop a specialized prompt tuning model for biomedical relation extraction, showing its effectiveness in few-shot learning [19]. El-Allaly et al. propose ADERel, an attentive joint model with a transformer-based weighted GCN for extracting adverse drug event (ADE) relations [20]. ADERel formulates the ADE RE task as an N-level sequence labeling, leveraging contextual and structural information [20]. The performance of traditional machine learning and deep learning models for relation extraction can be limited by several factors. One major challenge is accurately capturing complex relationships between entities, especially in context-dependent or inter-sentence scenarios, leading to reduced performance in tasks requiring high precision and recall [21]. Additionally, model performance can suffer from inadequate or noisy training data, affecting generalization [21,22]. Moreover, the reliance on large pre-trained language models such as Bidirectional Encoder Representations from Transformers (BERT) or Long Short-Term Memory (LSTM), can also pose challenges, as these models may not be well-suited for all RE tasks and may require extensive fine-tuning and adaptation [23].

Large language models (LLMs) have revolutionized natural language processing (NLP) by leveraging deep learning techniques to understand and generate human-like text [24]. Among these models, the Generative Pre-trained Transformer (GPT) stands out as a prominent example. Developed by OpenAI, GPT has gained widespread attention for its ability to perform a variety of language tasks, including text completion, translation, and summarization [25]. What sets GPT apart is its architecture, which is based on transformer neural networks [26]. This architecture allows GPT to effectively capture long-range dependencies in text, enabling it to generate coherent and contextually relevant responses [27]. Li et al. evaluated multiple pre-trained and fine-tuned LLMs on their ability to extract adverse events (AEs) using notes from the Vaccine Adverse Event Reporting System (VAERS), which has shown LLMs' capability in named entity recognition (NER) tasks [28]. Hu investigated the potential of ChatGPT, a large language model, for clinical NER in a zero-shot setting using two different prompt strategies [29]. However, few studies have explored the use of LLMs for extracting relations between entities, especially for a specific medical domain. To investigate the ability of LLMs in modeling acupuncture knowledge, we launched this study as an in-depth extension of our prior work in acupuncture knowledge extraction, which focused on extracting acupuncture point location entities based on the World Health Organization Standard Acupuncture Point Locations in the Western Pacific Region (WHO Standard) [30].

In this study, our objective is to leverage the power of GPT for relation extraction in the context of acupuncture points and human anatomy. Specifically, we aim to compare the performance differences between traditional deep learning models and LLMs, as well as between pre-trained LLMs and fine-tuned LLMs. This comparative analysis will provide insights into the effectiveness of using GPT and other models for RE tasks in the domain of acupuncture, and highlight the potential of LLMs to advance modeling of acupuncture knowledge.

## METHODS

Figure 1 shows the workflow of our study, which aimed to extract relations between acupuncture points using various models. We utilized data from the WHO Standard as the data source. We manually annotated and studied five types of relations. Five models were utilized: Bidirectional Encoder Representations from Transformers for Biomedical Text Mining (BioBERT), LSTM network, pre-trained GPT-3.5-turbo, fine-tuned GPT-3.5-turbo, and pre-trained GPT-4.

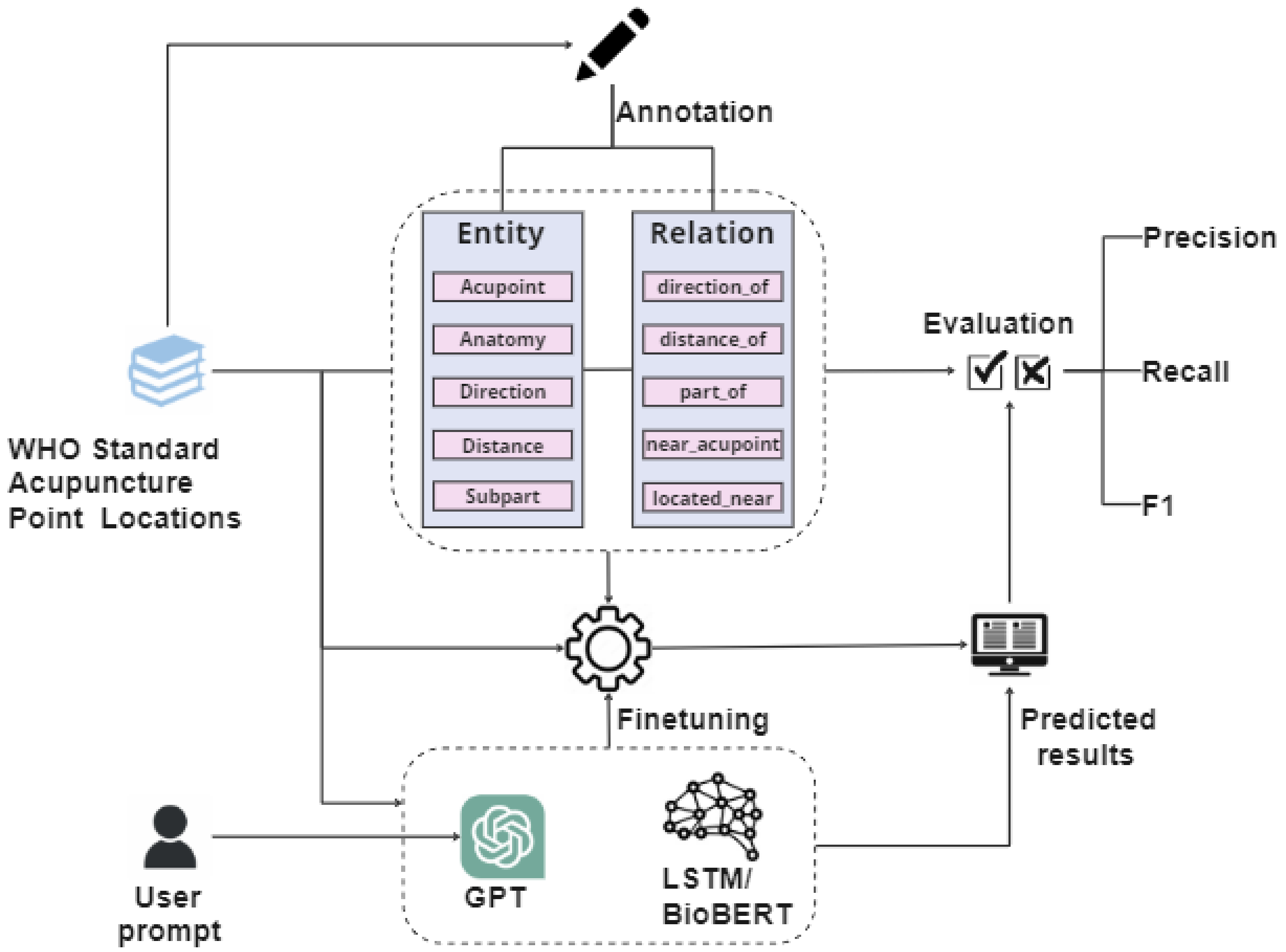

Figure 1 Overview of the framework

## Data Source

The WHO Standard is the data source we utilized in this study [31]. WHO Standard is a pivotal resource released in 2008, aiming to standardize the locations of acupuncture points [30,31]. WHO Standard elaborates on the locations of 361 acupoints across 14 systems and specifies the methodology to locate acupoints of the human body [30,31]. The book comprises two parts: General Guidelines for Acupuncture Point Locations and the WHO Standard Acupuncture Point Locations, offering essential terms, definitions, and measurement standards for acupoint locations [31]. The WHO Standard is expected to facilitate scientific communications in traditional medicine and contribute to evidence-based clinical efficacy for acupuncture therapy, making Eastern medicine more accessible and valuable in human healthcare [30,31].

## Annotation

In this study, we used the annotated entities in the studies carried out by Li et al. [30]. There are six types of entities: *Acupoints*, *Anatomy*, *Direction*, *Distance*, *General Location*, and *Subpart*. To precisely locate acupoints, we investigated five relation types: *direction_of* indicates the *Direction* of one relative *Acupoint* or *Anatomy* to the *Acupoint* of interest, *distance_of* denotes the *Distance* to the *Acupoint*/*Anatomy*, *part_of* represents the *Subpart* of the *Anatomy*, *near_acupoint* refers to adjacency to the relative *Acupoint*, and *located_near* signifies proximity of the *Anatomy* to the *Acupoint* of interest (shown in Table 1). The total annotated relation pairs are 3,174. All annotations were performed using CLAMP (version 1.6.6) [32].

In our annotation examples, acupoints were described using detailed spatial relationships, such as SP8 (Diji): 'On the tibial aspect of the leg, posterior to the medial border of the tibia, 3 B-cun inferior to SP9.' This narrative includes various key relationships, as illustrated in Table 1. For instance, the terms 'inferior to' and 'SP9' describe the acupoint's position below SP9, highlighting the 'direction_of' relationship. The distance between '3 B-cun' and 'SP9' is expressed as a 'distance of' relation, indicating that SP8 is located 3 B-cun inferior to SP9. Furthermore, relationships such as 'tibial aspect of' and 'leg,' and 'medial border of' and 'tibia,' provide additional context by specifying which part of the body the acupoint is located on and its proximity to specific anatomical landmarks, representing 'part_of' relationships. Additionally, the 'near_acupoint' relation is demonstrated by the proximity of 'SP8' and 'SP9,' indicating that these acupoints are close to each other. Lastly, the 'located_near' relation is exemplified by 'SP8' being near both the 'leg' and the 'tibia,' highlighting its close proximity to these anatomical structures.

Table 1 Types of relations annotated for SP8 (Diji) in WHO Standard

<table>
<tr><th>Relation Type</th><th>Definition</th><th>Example</th></tr>
<tr><td>direction_of</td><td>the Direction of one relative Acupoint or Anatomy to the Acupoint of interest</td><td>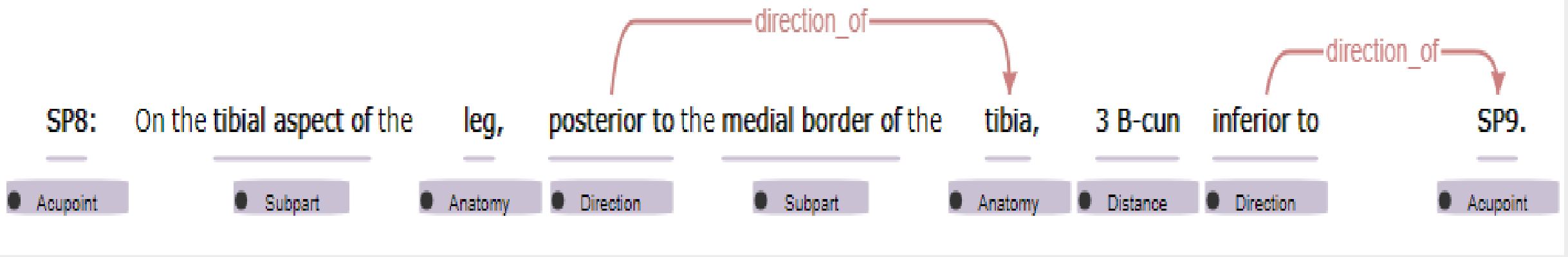</td></tr>
<tr><td>distance_of</td><td>the Distance to the Acupoint/Anatomy</td><td>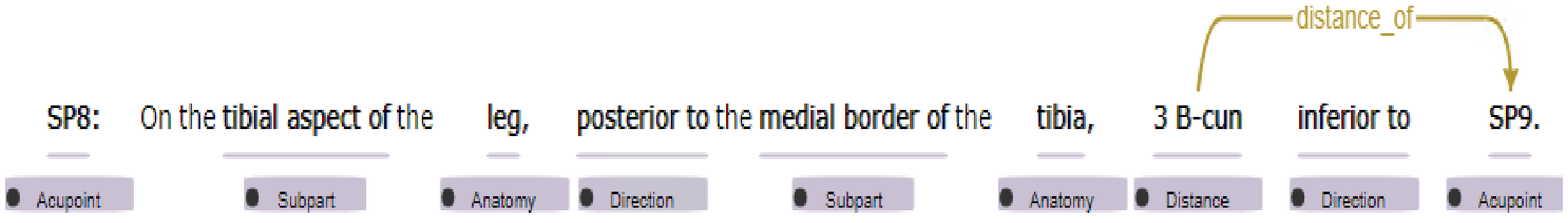</td></tr>
<tr><td>part_of</td><td>the Subpart of the Anatomy</td><td>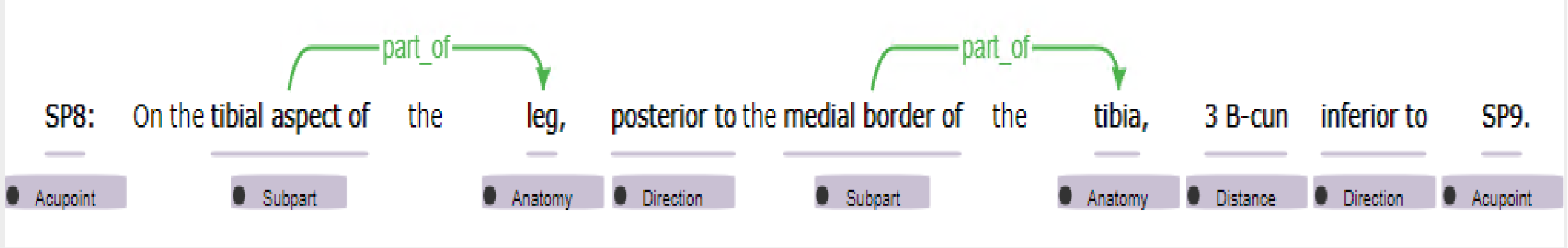</td></tr>
<tr><td>near_acupoint</td><td>adjacency to the relative Acupoint</td><td>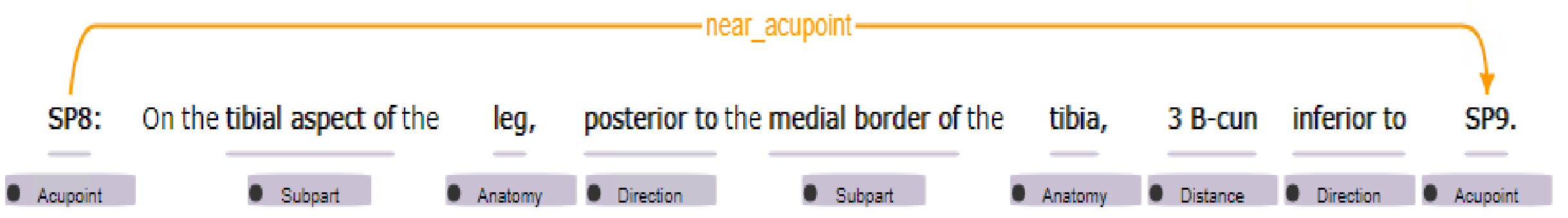</td></tr>
<tr><td>located_near</td><td>proximity of the Anatomy to the Acupoint of interest</td><td>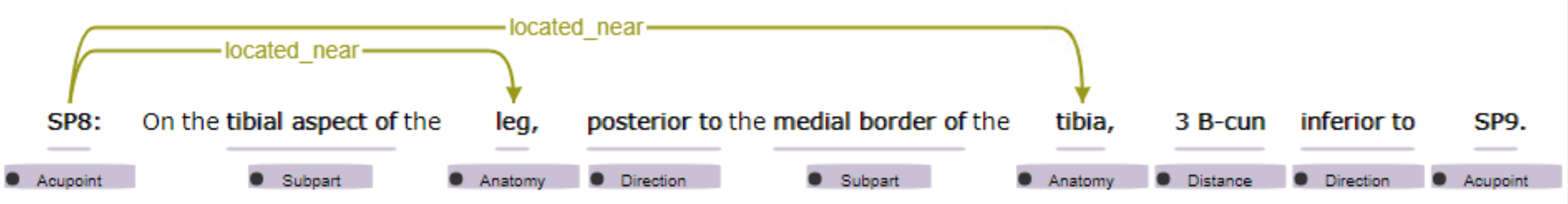</td></tr>
</table>

## Model

In this study, we explored the effectiveness of pre-trained GPT-3.5, fine-tuned GPT-3.5, pre-trained GPT-4, BioBERT, and LSTM models in extracting acupoint-based relations.

### GPT

GPTs are a class of deep learning models developed by OpenAI, based on the Transformer architecture [33]. These models are pre-trained on large text corpora to understand and generate human-like text [33,34]. Due to their ability to generate coherent and contextually relevant text, GPT models have been widely adopted for various NLP tasks including NER, question answering, machine translation, and text summarization [35].

### BioBERT

BioBERT is a specialized version of the BERT model, tailored for biomedical text [36]. Developed by the Korea University and Clova AI Research (NAVER Corp.), BioBERT is pre-trained on large-scale biomedical corpora, enabling it to better understand and process biomedical text [36]. This adaptation allows BioBERT to capture domain-specific features and terminologies, making it particularly effective for various tasks [36].

### LSTM

LSTM is a type of recurrent neural network (RNN) architecture that is designed to handle long-term dependencies in sequential data [37]. Unlike traditional RNNs, LSTM networks have a gating mechanism that allows them to selectively remember or forget information over time [38]. This capability makes LSTMs well-suited for tasks such as speech recognition, language modeling, and machine translation [39].

## Evaluation

In evaluating the performance of our model, we used our annotations as the gold standard. The micro-average approach calculates the total true positives, false positives, and false negatives between the gold standard and the predicted results across all classes and then computes precision, recall, and F1 score. The formulas for precision, recall, and F1 scores are shown below:

$$Precision = \frac{True\ positive}{True\ positive + False\ positive}$$

$$Recall = \frac{True\ positive}{True\ positive + False\ negative}$$

$$F-1 = \frac{2 \times Precision \times Recall}{Precision + Recall}$$

These metrics provide a comprehensive evaluation of the model's performance in relation extraction, capturing both its ability to identify relevant relations (recall) and its precision in correctly classifying them.

## Experiment Setup

In the experiment setup, the dataset consisting of 361 acupoints was randomly divided into training and test sets in an 8:2 ratio. This division ensured that 80% (N=288) of the acupoints were allocated to the training set for model training, while the remaining 20% (N=73) were reserved for evaluating the model's performance on unseen data in the test set.

In the experiment setup, we further analyzed the dataset to examine the distribution of each relation type within the training and test sets. This analysis provided insights into the balance of relation types and ensured that the model was trained and evaluated on a diverse set of relations, contributing to its robustness and generalization ability. The counts of each relation type in the training and test sets are summarized in Table 2 below.

Table 2 Relation statistics in the training and test set

| Relation type | Training | Test | Total |
|---|---|---|---|
| **direction_of** | 694 | 160 | 854 |
| **distance_of** | 238 | 66 | 304 |
| **part_of** | 467 | 107 | 574 |
| **near_acupoint** | 154 | 49 | 203 |
| **located_near** | 992 | 247 | 1,239 |

## GPT

In this study, we fine-tuned the GPT model for each relation type using specific prompts tailored to extract the corresponding relations. For instance, the prompt for the "direction_of" relation was formatted as follows:

> *"Textual information is below.\n---------------------\n{txt}\n---------------------\nEntity information is below(Format is:'T(id)\t(entity_type) (start_offset) (end_offset)'\t'(entity_text)'){entities}\nQuery:Now we have to conduct the relation extraction task. We want to extract 'direction_of' first. 'direction_of' refers to the direction of one relative 'Acupoint'or 'Anatomy' to the 'Acupoint'. The starting entity type of 'direction_of' is 'Direction' entity and the end entity of 'direction_of' is the relative 'Acupoint' or 'Anatomy' entity."*

In Figure 2, we provided a training example for the "direction_of" relation extraction using the acupoint BL12 as an illustrative case. This approach allowed us to fine-tune the GPT model for each relation type, enhancing its ability to extract specific relations accurately and efficiently from textual knowledge.

In Table 3, we present the key parameters used for the GPT models in our study. For all GPT models, including the pre-trained GPT-3.5, fine-tuned GPT-3.5 and pre-trained GPT-4, we set the temperature parameter to 0.3 and the maximum number of tokens to 4,096.

In the case of fine-tuned GPT-3.5, the number of epochs (value=2), batch size (value=1), and learning rate multiplier (value=1) were used during the fine-tuning process. These parameters were chosen to balance the model's performance and computational efficiency, ensuring effective training and inference for RE tasks.

Table 3 Parameters for GPT models

| **Parameter** | **Value** |
|---|---|
| **Temperature** | 0.3 |
| **Max token** | 4,096 |
| **n_epochs**[a] | 2 |
| **batch_size**[a] | 1 |
| **learning_rate_multiplier**[a] | 1 |

[a]Fine-tuned GPT 3.5 only

Prompt

```
"role": "system",
"content": "You are a relation retrieval expert that is good at extract
            relations between entities. To facilitate output extraction, please output the
            relation directly and format the relation as
           "Relation sequence number(Starting from R1, R2,...) direction_of Arg1:Entity
            sequence number Arg2:Entity sequence number"
           from the following text. If no relations, please do not output anything."

"role": "user",
"content": "Textual information is below.
            {BL12: In the upper back region, at the same level as the inferior border of the
            spinous process of the second thoracic vertebra (T2), 1.5 B-cun lateral to the
            posterior median line. }
            Entity information is below (Format is:'T(id) (entity_type) (start_offset)
            (end_offset)(entity_text)')
           {T1      Acupoint 0 4         BL12
            T2      Anatomy 13 30        upper back region
            T3      Direction 39 52      same level as
            T4      Subpart 57 98        inferior border of the spinous process of
            T5      Anatomy 103 132      second thoracic vertebra (T2)
            T6      Distance 134 143     1.5 B-cun
            T7      Direction 144 154    lateral to
            T8      Anatomy 159 180      posterior median line}
            Query: Now we have to conduct the relation extraction task. We want to extract
            'direction_of' first. 'direction_of' refers to the direction of one relative
            'Acupoint'or 'Anatomy' to the 'Acupoint'. The starting entity type of
            'direction_of' is 'Direction' entity and the end entity of 'direction_of' is the
            relative 'Acupoint' or 'Anatomy' entity."
```

Annotations

```
"role": "assistant",
"content": "R1     direction_of Arg1:T7 Arg2:T8
            R2     direction_of Arg1:T3 Arg2:T5"
```

Figure 2 One example of training set for “direction_of” relation of BL12 (Fengmen)

### LSTM, BioBERT

For the parameters (shown in Table S1), BioBERT v1.1 was utilized as the version for BioBERT in the baseline comparison. Both LSTM and BioBERT were trained with a batch size of 32. For LSTM, training extended over 100 epochs with a learning rate of 1e-3, whereas BioBERT was trained for 3 epochs with a learning rate of 5e-5. Furthermore, the maximum number of training steps for LSTM was set at 14,400, while the maximum sequence length for BioBERT was restricted to 128.

Fine-tuning and inference for the pre-trained GPT models were conducted on a server equipped with 8 Nvidia A100 GPUs, each with a memory capacity of 80GB. In contrast, the LSTM and BioBERT models were fine-tuned and evaluated on a server with 5 Nvidia V100 GPUs, each providing a memory capacity of 32GB.

## RESULTS

Based on the precision results presented in Figure 3(a), we observed notable performance differences among the models across different relation types in terms of precision. For the 'direction_of' relation, the fine-tuned GPT 3.5 model achieved the highest precision of 0.96, indicating its strong capability in accurately identifying the direction of one relative acupoint or anatomy to the acupoint of interest. Similarly, for the 'distance_of' relation, the fine-tuned GPT 3.5 model also performed exceptionally well with a precision of 0.88. In contrast, the BioBERT model showed comparable precision scores across all relation types except “direction_of”. Overall, the fine-tuned GPT 3.5 model demonstrated the highest micro-average precision of 0.91.

The recall results, as shown in Figure 3(b), reveal interesting insights into the performance of different models for relation extraction in acupuncture. The fine-tuned GPT 3.5 model achieved the highest recall for most relation types, particularly excelling in the 'direction_of' and 'part_of' relations with recalls of 0.99 and 0.96, respectively. This indicates its strong ability to correctly identify these relationships within the text. However, the pre-trained GPT-4 model showed lower recall scores across all relation types, suggesting that it may struggle to capture the nuances of acupoint-related relations.

Overall, the fine-tuned GPT 3.5 model demonstrated the highest micro-average recall of 0.94.

The F1 scores presented in Figure 3(c) provide a comprehensive view of the overall performance of the models in relation extraction for acupuncture. Notably, the fine-tuned GPT 3.5 model achieved the highest F1 scores across most relation types, with particularly strong performance in the 'direction_of' and 'part_of' relations, achieving F1 scores of 0.97 and 0.91, respectively. This highlights the effectiveness of fine-tuning in improving the model's ability to extract complex relations in acupuncture text. In contrast, the pre-trained GPT-4 model showed lower F1 scores, indicating that it may struggle to generalize to acupoint-related relations. Overall, the fine-tuned GPT 3.5 model demonstrated the highest micro-average F1 score of 0.92, underscoring its effectiveness in relation extraction for acupuncture.

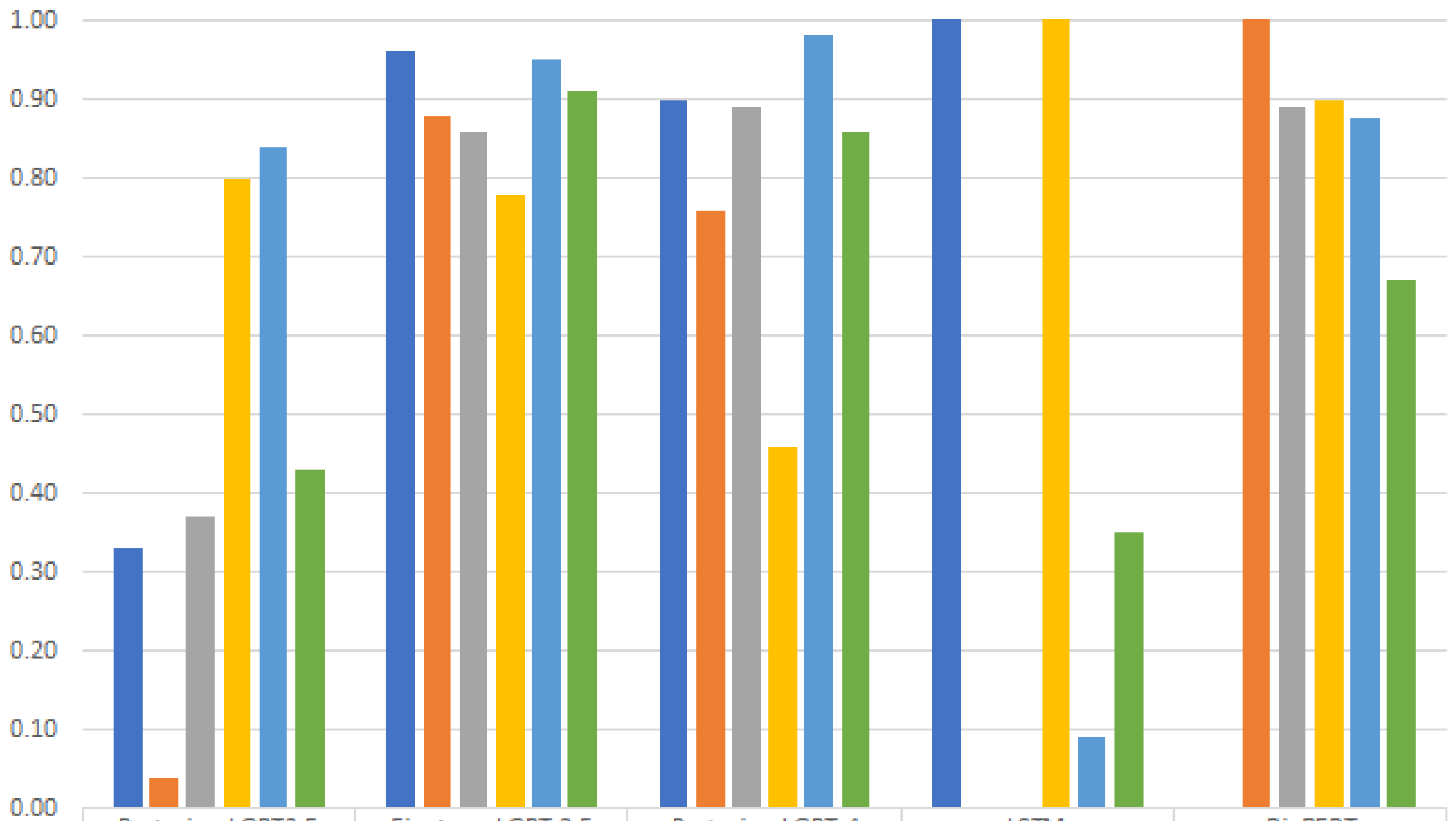

| | Pretrained GPT3.5 | Finetuned GPT 3.5 | Pretrained GPT-4 | LSTM | BioBERT |
|---|---|---|---|---|---|
| direction_of | 0.33 | 0.96 | 0.9 | 1.00 | 0.00 |
| distance_of | 0.04 | 0.88 | 0.76 | 0.00 | 1.00 |
| part_of | 0.37 | 0.86 | 0.89 | 0.00 | 0.89 |
| near_acupoint | 0.80 | 0.78 | 0.46 | 1.00 | 0.90 |
| located_near | 0.84 | 0.95 | 0.98 | 0.09 | 0.88 |
| Micro-average | 0.43 | 0.91 | 0.86 | 0.35 | 0.67 |

direction_of distance_of part_of near_acupoint located_near Micro-average

(a) precision

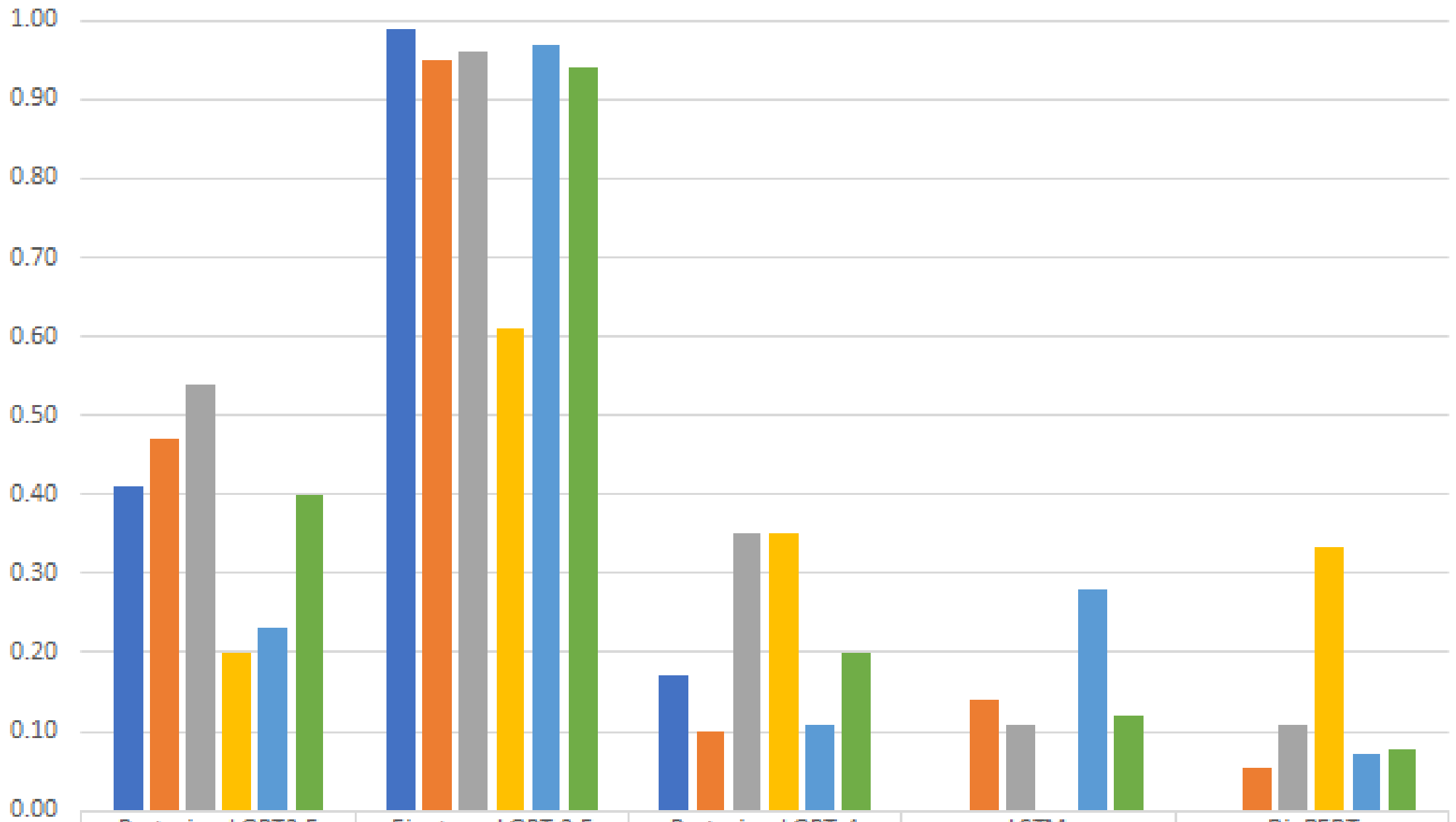

| | Pretrained GPT3.5 | Finetuned GPT 3.5 | Pretrained GPT-4 | LSTM | BioBERT |
|---|---|---|---|---|---|
| direction_of | 0.41 | 0.99 | 0.17 | 0.00 | 0.00 |
| distance_of | 0.47 | 0.95 | 0.10 | 0.14 | 0.05 |
| part_of | 0.54 | 0.96 | 0.35 | 0.11 | 0.11 |
| near_acupoint | 0.20 | 0.61 | 0.35 | 0.00 | 0.33 |
| located_near | 0.23 | 0.97 | 0.11 | 0.28 | 0.07 |
| Micro-average | 0.40 | 0.94 | 0.20 | 0.12 | 0.08 |

direction_of  distance_of  part_of  near_acupoint  located_near  Micro-average

(b) recall

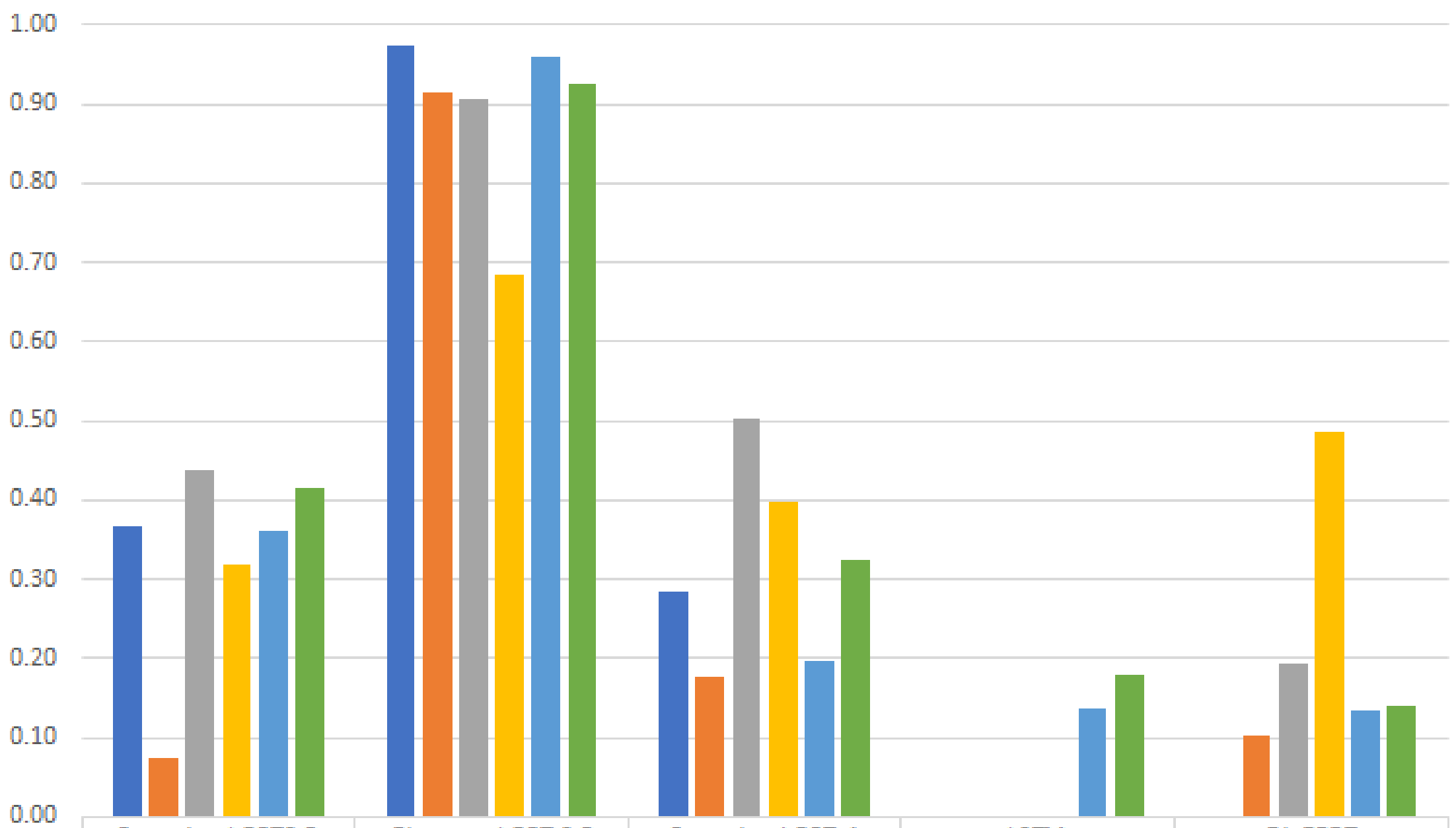

| | Pretrained GPT3.5 | Finetuned GPT 3.5 | Pretrained GPT-4 | LSTM | BioBERT |
|---|---|---|---|---|---|
| direction_of | 0.37 | 0.97 | 0.29 | 0.00 | 0.00 |
| distance_of | 0.07 | 0.91 | 0.18 | 0.00 | 0.10 |
| part_of | 0.44 | 0.91 | 0.50 | 0.00 | 0.19 |
| near_acupoint | 0.32 | 0.68 | 0.40 | 0.00 | 0.49 |
| located_near | 0.36 | 0.96 | 0.20 | 0.14 | 0.13 |
| Micro-average | 0.41 | 0.92 | 0.32 | 0.18 | 0.14 |

(c) F1

Figure 3 Performance on each relation type by different models

# DISCUSSION

The proposed study provides a comprehensive analysis of RE tasks. Our study compared the performance of traditional deep learning models (LSTM, BioBERT) with GPT in RE tasks related to acupoint locations. We also examined the performance difference between pre-trained GPT models and fine-tuned GPT models.

Our results indicate that the fine-tuned GPT-3.5 model outperformed both LSTM and BioBERT across all evaluated relation types. Specifically, the fine-tuned GPT-3.5 model achieved the highest F1 scores, demonstrating its effectiveness in accurately extracting relations related to acupoints. Furthermore, we observed that pre-trained GPT models, while showing strong performance in general language tasks, did not perform as well as fine-tuned GPT models in relation extraction for acupuncture point locations. This suggests that fine-tuning the GPT model on domain-specific data significantly improves its ability to extract complex body location relations related to acupoints.

The performance difference between traditional deep learning models and LLMs like GPT can be attributed to several factors. LLMs, with their transformer architecture, are better able to capture long-range dependencies in text, allowing them to generate more coherent and contextually relevant responses. While BioBERT also utilizes transformer architecture, its performance in this study may be inferior to GPT due to differences in pre-training data and fine-tuning strategies, which can significantly impact model performance.

Traditional deep learning models often struggle with capturing long-range dependencies in text, which is crucial for understanding context and extracting nuanced relations. Additionally, these models may require extensive fine-tuning and adaptation to perform well on specific tasks, making them less versatile compared to LLMs. Moreover, the interpretability of traditional deep learning models is often limited, posing challenges to understand the reasoning behind their predictions. These factors contribute to reduced performance in tasks requiring high precision and recall, especially in context-dependent scenarios.

Additionally, fine-tuning the GPT model on domain-specific data enables it to learn the nuances and complexities of relation extraction in acupuncture, leading to improved performance compared to models that are not fine-tuned. Fine-tuning allows the model to adapt its weights and parameters to the specific characteristics of the acupuncture RE task, enhancing its ability to extract accurate and contextually relevant relations.

Our study contributes to the field of healthcare informatics, as well as traditional and complementary medicine, in several significant ways. Firstly, we leverage the power of LLMs, specifically the GPT, for RE tasks in the domain of acupuncture. This application of advanced NLP techniques to traditional and complementary medicine demonstrates the potential of integrating modern technology with ancient acupuncture practices to improve the acupuncture outcome. Secondly, we provide a comparative analysis of the performance differences between traditional deep learning models and LLMs, as well as between pre-trained LLMs and fine-tuned LLMs in the domain of information extraction. This comparative study not only sheds light on the effectiveness of using LLMs like GPT for RE tasks but also provides insights into the benefits of fine-tuning these models for specific domains, such as acupuncture. Thirdly, our study highlights the challenges and opportunities in using NLP techniques for extracting complex relationships from textual acupuncture knowledge. By examining the limitations of existing models and proposing solutions to improve their performance, we contribute to advancing the state-of-the-art in NLP for acupuncture applications.

One of the key strengths of our study is the utilization of the WHO Standard as our corpus. This widely accepted resource provides a formal and standardized conceptual framework for acupuncture point locations, ensuring the accuracy and reliability of our results. By using this authoritative source, our findings can be directly applied to real clinical settings, enhancing the practical relevance and utility of our study in acupuncture practice. Additionally, the WHO Standard's comprehensive coverage of acupoint locations across different systems and its detailed methodology for acupoint locations ensure the generalizability of our findings. Additionally, our study highlights the potential of fine-tuning LLMs for relation extraction in traditional and complementary medicine domains, showcasing their adaptability and versatility.

Limitations of this study include the use of a specific dataset with limited size and format that may not fully represent the diversity of textual knowledge related to acupuncture, and the challenges of manual annotation prone to error and bias. Future work would focus on domain adaptation and acupuncture-related knowledge integration to improve model performance in textual knowledge.

## Error Analysis

We conducted an error analysis of the fine-tuned GPT-3.5 model's performance in relation identification, as shown in Table 4. The analysis revealed varying error rates across different relation types, with notable findings in the "near_acupoint" relation. This relation type exhibited a high false positive rate of 32.5% and a relatively high false negative rate of 26.53%. These results suggest that the model struggles with accurately identifying acupoints that are close to each other, possibly due to the complexity of spatial relationships and the nuanced context required for such distinctions.

Additionally, all classes showed high positive rates compared to the other two error types—false negative and incorrect relation types. Despite these challenges, the model demonstrated a low error rate in assigning incorrect entity types, indicating a strong understanding of entity types in the context of acupoint descriptions. These findings highlight the model's strengths and weaknesses in relation identification, pointing to areas for improvement, particularly in spatial relation understanding.

Table 4 Error Analysis for fine-tuned GPT-3.5

| | **False Positive (out of machine annotated entities)** | **False Negative (out of human annotated entities)** | **Incorrect Relation Type (out of machine annotated entities)** |
|---|---|---|---|
| **direction_of** | 11/168, 6.55 % | 1/160, 0.63% | 0/168, 0% |
| **distance_of** | 13/75, 17.33% | 1/66, 1.52% | 0/75, 0% |
| **part_of** | 16/119, 13.45% | 1/107, 0.93% | 0/119, 0% |
| **near_acupoint** | 13/40, 32.5% | 13/49, 26.53% | 0/40, 0% |
| **located_near** | 13/252, 5.16% | 4/247,1.62% | 1/252, 0.4% |

Based on the detailed error analysis of the fine-tuned GPT-3.5 model in relation extraction for acupoint descriptions, several key observations and patterns were uncovered. The model occasionally fails to accurately identify adjacent acupoints, as seen in the description of ST38 (shown in Figure 4) where it incorrectly identifies ST35 as the acupoint of interest instead of correctly identifying the "near_acupoint" relationship between ST38 and ST35 or ST40. This indicates a difficulty in understanding complex spatial relationships in complex descriptions.

The model also struggles with identifying relationships that span across sentences, as seen in the description of ST31 ("St31 is located at the deepest point in the depression inferior to the apex of this triangle. Note 2: ST31 is located at the intersection of the line connecting the lateral end of the base of the patella with the anterior superior iliac spine, and the horizontal line of the inferior border of the pubic symphysis.”) where it incorrectly identifies a "direction_of" relationship between "inferior to" and "anterior superior iliac spine." This error may stem from the model's inability to effectively track and integrate information across multiple sentences. It may lack the contextual understanding which causes it to recognize that "inferior to" relates to “the anterior superior iliac spine”. Additionally, in the description of LR10 (“LR10: On the medial aspect of the thigh, 3 B-cun distal to St30, over the artery.”), the model incorrectly identifies a "part_of" relationship between "medial aspect of" and "thigh." This error suggests that the model may struggle with accurately identifying anatomical relationships between acupoints and body parts, possibly due to a lack of profound understanding of anatomical structures and their spatial relationships.

While less frequent than false positives, false negatives also occur, as seen in the description of BL38 (“BL38: On the posterior aspect of the knee, just medial to the biceps femoris tendon, 1 B-cun proximal to the popliteal crease. Note: With the knee in slight flexion, BL38 is located medial to the biceps femoris tendon, 1 B-cun proximal to BL39.”) where the model fails to detect the "located_near" relationship between "BL38" or "BL39" and "knee." This indicates a limitation in the model's ability to capture subtle spatial relationships. The errors may be due to a lack of comprehensive knowledge about acupoints, preventing it from integrating acupoint information with anatomical knowledge effectively. Additionally, the model may not have been trained on a sufficiently large corpus of acupoint-related data, limiting its ability to learn complex relationships. Relations between anatomical structures and acupoints are inherently complex and may be challenging for the model to detect accurately.

These errors highlight limitations in the model's ability to contextualize information over multiple sentences and to accurately interpret complex anatomical relationships. To address these issues, it may be beneficial to expand the model's knowledge base on acupoints, expand its training data with a larger and more diverse corpus, and enhance its ability to comprehend context over multiple sentences. Improving the model's contextual understanding and strengthening its knowledge of anatomical structures and their relationships could also help mitigate these errors and elevate its performance in relation extraction for acupoint descriptions, upgrading its accuracy and reliability in this domain.

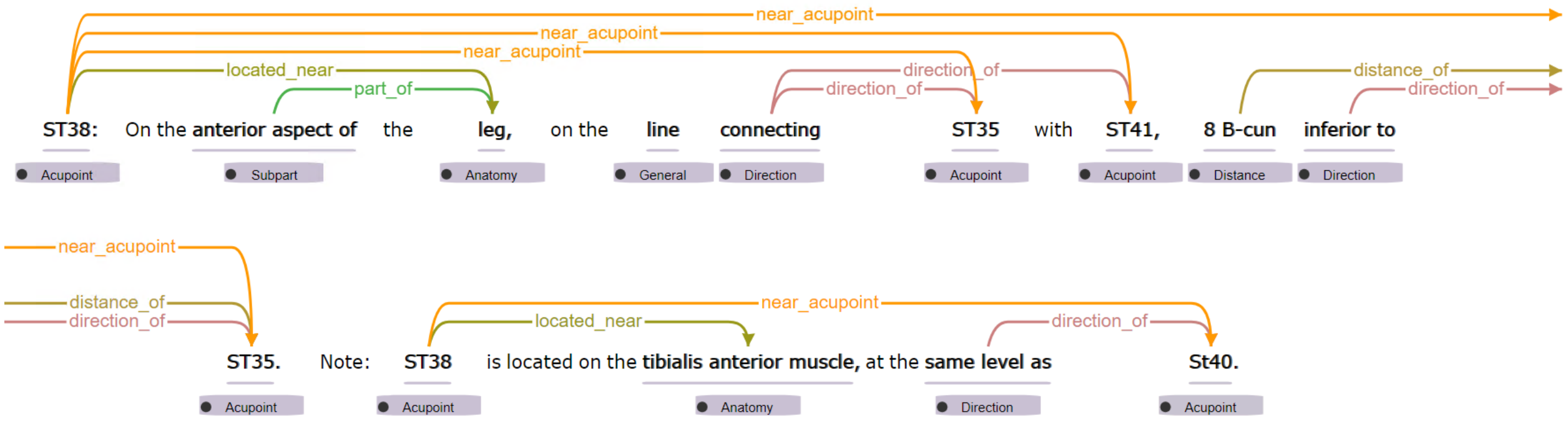

(a) gold standard

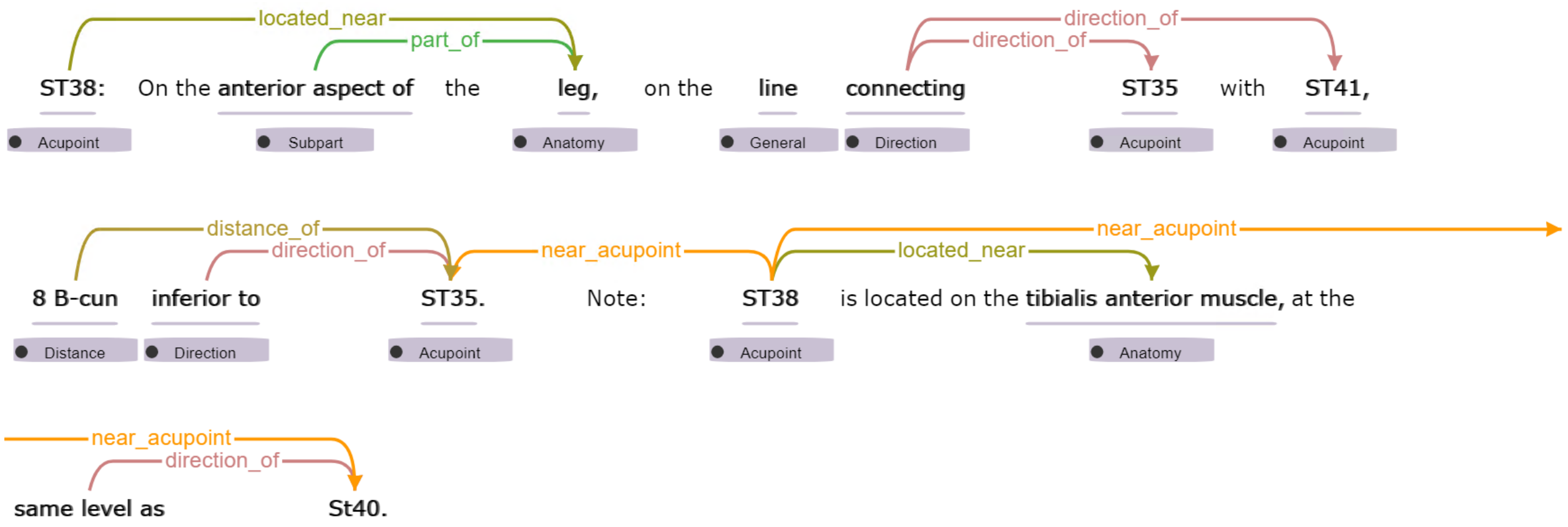

(b) relation predictions from fine-tuned GPT-3.5

Figure 4 Error analysis for ST38 (Tiaokou)

# CONCLUSION

This study underscores the effectiveness of LLMs like GPT in extracting relations related to acupoint locations, highlighting their value in the acupuncture domain. By utilizing LLMs, we have demonstrated the potential for accurate modeling of acupoint location knowledge and further promoting precise locating of the acupoints during acupuncture practice. The findings also contribute to advancing informatics applications in traditional and complementary medicine, showcasing the potential of LLMs in natural language processing.

## Ethics Approval and Consent to Participate

Not applicable.

## Competing Interests Statement

The authors have no competing interests to declare.

## Contributorship Statement

YL and XP designed the study. YL developed the pipeline. YL and JL built the model, and YL performed visualization. CT and XZ offered technical resource support. DP contributed to data collection. NH and SP participated in data annotation. YL drafted the manuscript. HX and NH supervised the study, and NH critically revised the manuscript.

## Data Availability Statement

The data underlying this article will be shared on reasonable request to the corresponding author.

## Code Availability Statement

The code underlying this article will be shared on reasonable request to the corresponding author.

## Abbreviations

| | |
|---|---|
| ADE | adverse drug event |

AE adverse event

BERT Bidirectional Encoder Representations from Transformers

BioBERT Bidirectional Encoder Representations from Transformers for Biomedical Text Mining (BioBERT)

GPT Generative Pre-trained Transformer

LLM large language model

LSTM Long Short-Term Memory

NER named entity recognition

NLP natural language processing

RE relation extraction

RNN recurrent neural network

VAERS Vaccine Adverse Event Reporting System

WHO Standard World Health Organization Standard Acupuncture Point Locations in the Western Pacific Region

# SUPPLEMENTARY MATERIAL

Table S1. Parameters for LSTM and BioBERT

| **Parameters** | **LSTM** | **BioBERT(BioBERT v1.1)** |
|---|---|---|
| **Batch size** | 32 | 32 |
| **epochs** | 100 | 3 |
| **Learning rate** | 1e-3 | 5e-5 |
| **Maximum training steps** | 14,400 | / |
| **Maximum sequence length** | / | 128 |